\documentclass[letterpaper, 10 pt, conference]{ieeeconf} 

\IEEEoverridecommandlockouts                              

\usepackage{graphicx}
\usepackage{amsmath}
\usepackage{xcolor}
\usepackage{bm}
\usepackage[T1]{fontenc}

\makeatletter
\let\NAT@parse\undefined
\makeatother
\usepackage[colorlinks,citecolor=blue,urlcolor=blue, linkcolor=blue, bookmarks=false,hypertexnames=true]{hyperref}

\title{\LARGE \bf 
Effective Virtual Reality Teleoperation of an Upper-body Humanoid with Modified Task Jacobians and  Relaxed Barrier Functions for Self-Collision Avoidance}
\author{Steven Jens Jorgensen and  Ravi Bhadeshiya \\
\{stevenjorgensen, ravibhadeshiya\}@apptronik.com \\
Apptronik Inc., Austin TX, USA}

\begin{document}

\maketitle
\thispagestyle{empty}
\pagestyle{empty}

\begin{abstract}
We present an approach for retartgeting off-the-shelf Virtual Reality (VR) trackers to effectively teleoperate an upper-body humanoid while ensuring self-collision-free motions. Key to the effectiveness was the proper assignment of trackers to joint sets via modified task Jacobians and relaxed barrier functions for self-collision avoidance. The approach was validated on Apptronik's Astro hardware by demonstrating manipulation capabilities on a table-top environment with pick-and-place box packing and a two-handed box pick up and handover task.
\end{abstract}


\section{INTRODUCTION}
Despite advances in robot autonomy, teleoperation \cite{stotko2019vr, peppoloni2015immersive, naceri2021vicarios}
remains a practical approach for remote surveying and intervention \cite{krotkov2017darpa, jorgensen2019deploying, rouvcek2019darpa}. While direct teleoperation is not viable with long network latencies, it remains a useful tool  for human-to-robot imitation learning \cite{zhang2018deep, mandlekar2020human} which will enable future robots to be more autonomous.

A core problem with direct teleoperation is retargeting human-to-robot movements, which is an active area of research
\cite{fernando2012design, penco2019multimode, darvish2019whole, elobaid2019telexistence, ishiguro2020bilateral,  ihmcicra2021workshop}. 
In a previous work, three 6 degree-of-freedom (DoF) trackers comprising of a VR headset and two controllers were used to fully control the pelvis height, torso, arms, and head of the NASA Valkyrie humanoid \cite{ihmcicra2021workshop, jorgensen2022cockpit}. With this approach, since there are more joints than tracker DoFs ($n_j > n_t$), multiple solutions for retargeting exist. Redundancy resolution is done by adding biasing posture tasks \cite{elobaid2019telexistence} and appropriate weighting of end-effector pose tasks \cite{ihmcrobotics2021}.  Another approach is to add more trackers to the operator with a full-body suit \cite{darvish2019whole}, here there are more tracker DoFs than robot joints ($n_t > n_j$). In either case, some form of weight tuning of tasks is required to obtain a desired retargeted behavior. An appropriate weight set can be difficult to identify and in some cases poor tuning of these weights can cause unwanted behaviors such as oscillations \cite{elobaid2019telexistence}.

\begin{figure}
\centerline{\includegraphics[width=1.0\columnwidth]{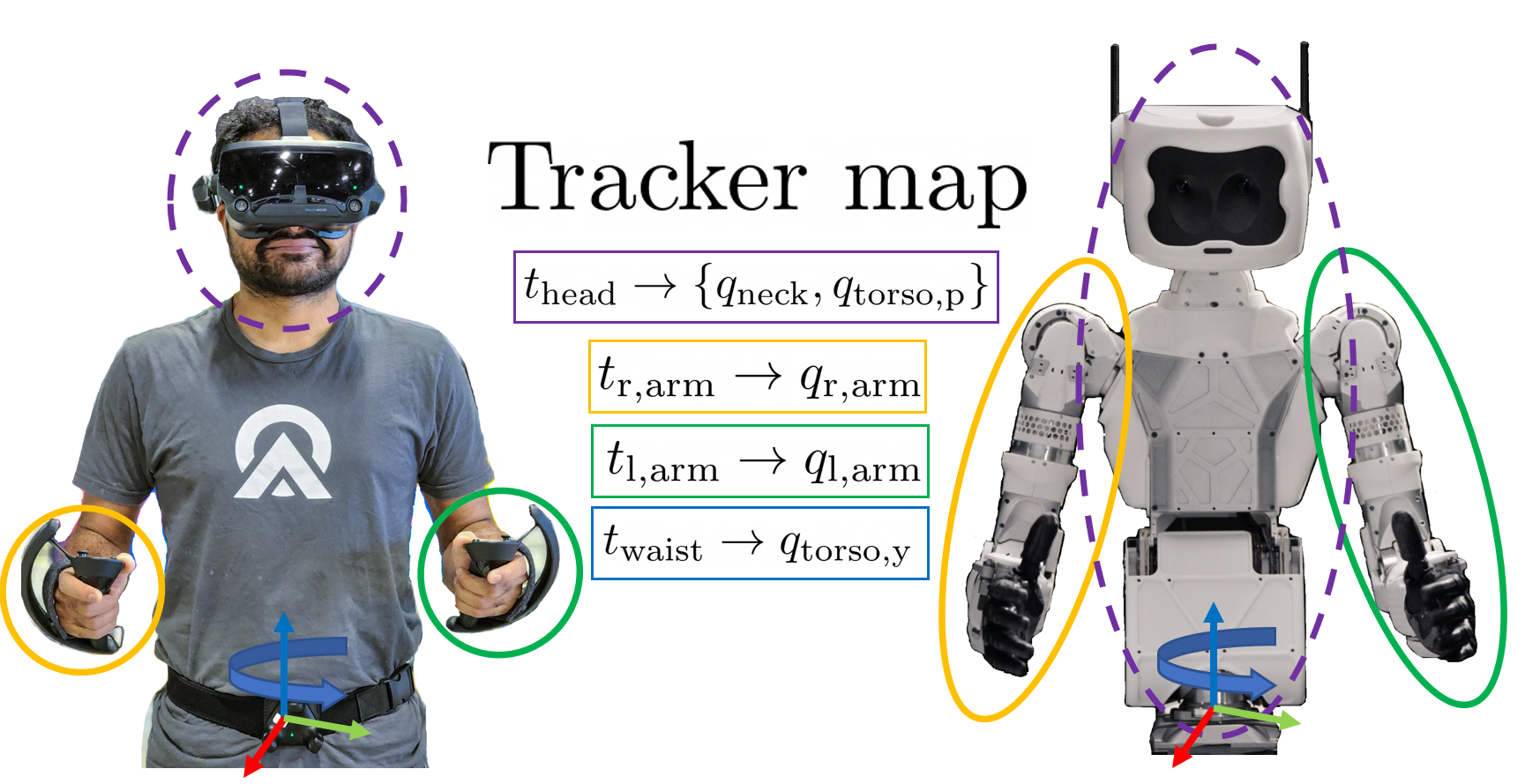}}
\caption{Four 6 DoF trackers (headset, left controller, right controller, waist tracker) and the set of joints they control on the robot. Using modified task Jacobians, the headset orientation controls the neck joints, the hand controllers' pose control only the arm joints, the waist tracker's vertical axis controls the torso yaw joint, and the forward position of the headset controls the leaning angle of the robot using the torso pitch joint.}
\label{fig:tracker-map} 
\end{figure}

In contrast, we propose to utilize a minimum set of trackers ($n_j = n_t$) and assign only a set of joints for each tracker DoF by modifying the corresponding task Jacobian for the retargetting task. This minimizes the responsibility of each joint, removes operational space task conflicts, conditions the retargeting behavior, and also informs the operator apriori which trackers map to which joints  (Fig.~\ref{fig:tracker-map}). This approach of proper task allocation was effective in certain bipedal locomotion approaches \cite{gong2019feedback, kim2020dynamic} and appears to be effective for teleoperation as well.

Finally, during direct teleoperation, it can be burdening and unsafe for the operator to also consider robot-self collisions on top of commanding the robot as part of regular operations.  An easy approach is to reject joint commands that would cause the robot to self-collide using a collision library checker \cite{pan2012fcl}. However, this tends to cause abrupt pauses when performing a task. A better approach is to include self-collision avoidance as part of the Inverse-Kinematics (IK) problem of retargeting. To our knowledge, most published works ignore the self-collision avoidance problem and rely on the operator to execute safe behaviors. The VR interface for the NASA Valkyrie robot \cite{ihmcrobotics2021, jorgensen2022cockpit} is an exception as it uses repulsive potential fields. We propose that  signed-distance and relaxed-barrier functions are an improved approach to handle self-collisions. 

\section{APPROACH OVERVIEW}
\textbf{Modified Task Jacobians} For a given operational task, $\bm{x}$, such as a robot end-effector pose goal, a Jacobian, $\bm{J}(\bm{q})$ relating the task velocities to the robot's joint velocities can be obtained from the current joint state, $\bm{q}$, of the robot. The columns of the Jacobian describe a joint's contribution to the incremental change in task coordinates \cite{lynch2017modern}. For instance, let us define the joint state vector $\bm{q} = [\bm{q}_{\text{torso,p}}, \bm{q}_{\text{torso,y}}, \bm{q}_{\text{neck}},  \bm{q}_{\text{l,arm}}, \bm{q}_{\text{r,arm}}]^T$, where $\bm{q}_{\text{l,arm}}$ for example is the set of joints corresponding to the robot's left arm. Then, the Jacobian of a task can be represented as follows, 
\begin{align}
    \dot{\bm{x}} &= \bm{J}(\bm{q}) \dot{\bm{q}} \label{eq:xdot}\\
     &= \Bigg[\frac{\partial\bm{x}}{\partial\bm{q}_{\text{torso,p}}},
     \frac{\partial\bm{x}}{\partial\bm{q}_{\text{torso,y}}},
     \frac{\partial\bm{x}}{\partial\bm{q}_{\text{neck}}},
     \frac{\partial\bm{x}}{\partial\bm{q}_{\text{l,arm}}},
     \frac{\partial\bm{x}}{\partial\bm{q}_{\text{r,arm}}}\Bigg] \dot{\bm{q}} \nonumber
\end{align}
The proposed \textit{modified task Jacobian} approach  removes unwanted joint contributions\footnote{Mathematically, this can be done by post-multiplying the Jacobian with a selection matrix, $\bm{S}$, i.e. $\bm{J}_m = \bm{J}\bm{S}$}. For example, when mapping the user's left hand tracker to the robot's left hand, we modify the task Jacobian so that only left arm joints are used for this retargeting task and ignore the torso's joint contributions to the velocities of the left hand, namely,
\begin{align}
    \dot{\bm{x}}_{\text{l,hand}} = \Bigg[{0, 0, 0,
     \frac{\partial\bm{x}}{\partial\bm{q}_{\text{l,arm}}},
     0}\Bigg] \dot{\bm{q}}.
\end{align}
This decomposition of joint responsibility makes the robot's behavior predictable to the operator as the mapping between each tracker to a joint set is clear. 

\textbf{Relaxed Barrier Functions for Self-Collision Avoidance} 
Collision avoidance has been traditionally incorporated to the IK problem as repulsive potential fields \cite{ihmcrobotics2021}. However, potential fields suffer from local minima in the presence of clutter and can cause unwanted behavior \cite{khatib1986real}. An alternative is to use signed-distance  constraints between convex shapes \cite{schulman2014motion}. Recently, signed-distance functions are enforced in real-time using control barrier functions as part of the inequality constraint \cite{khazoom2022humanoid} or as a soft constraint \cite{chiu2022collision} using relaxed barrier functions \cite{feller2016relaxed, grandia2019feedback}. We propose to use soft constraints for computational reasons: first, as a soft constraint, best-effort solutions are preferred over optimization infeasibility, second the inequality evaluation step is skipped on most Quadratic Programming (QP) based solvers \cite{goldfarb1983numerically, ferreau2014qpoases}, which improves solve time speed and consistency.

\textbf{Robust IK with Relaxed Barrier Functions}
For a given list of $N_t$ operational tasks (e.g. end-effector poses) and $N_c$ collision pairs, joint velocity solutions are found 
using a QP-based IK solver\footnote{\url{https://github.com/stephane-caron/pymanoid}} \cite{caron2016thesis} with the following form,
\begin{align}
\underset{\dot{\bm{q}} }{\text{min}} \sum_{i=1}^{N_t} w_i ||\bm{K}_{p,i} \bm{e}_i - \bm{J}_i\dot{\bm{q}}|| +\dot{\bm{q}}^T \bm{W}_i \dot{\bm{q}} + \sum_{j=1}^{N_c} \tilde{B}_j(h_j(\bm{q}), \dot{\bm{q}}).
\end{align}
The first term is a weighted least-squares solution of Eq.~\ref{eq:xdot}
with gain $\bm{K}_{p,i}$ and task error $\bm{e}_i$. The second term is an adaptive regularization matrix that ensures robust numerical solutions are available even if $\bm{J}$ is ill-conditioned \cite{sugihara2011solvability}. The last term is a quadratic approximation of the relaxed barrier function, $B(\cdot)$, where $h_j(\bm{q})$ is a signed distance. When expanded, $\tilde{B}_j(h_j(\bm{q}), \dot{\bm{q}})$ has the following form
\begin{align}
    \tilde{B}(h_j(\bm{q}), \dot{\bm{q}}) &=  B(h_o) +  \bigg(\frac{\partial B}{\partial h} \frac{\partial h}{\partial \bm{q}}\bigg)^T \dot{\bm{q}} \ + \\
    &  \frac{1}{2} \ \dot{\bm{q}}^T \bigg(\frac{\partial h}{\partial \bm{q}}^T \frac{\partial^2 B}{\partial h^2} \frac{\partial h}{\partial \bm{q}}\bigg) \dot{\bm{q}} \nonumber
\end{align}



\begin{figure}
\centerline{\includegraphics[width=1.0\columnwidth]{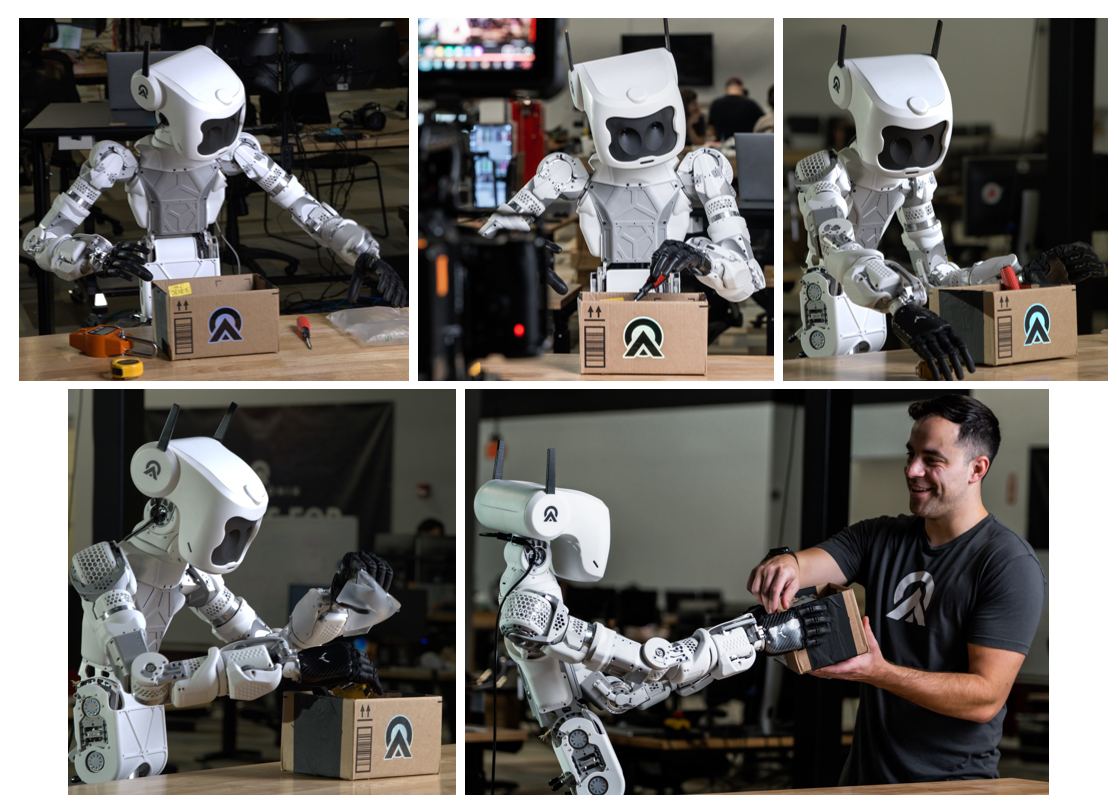}}
\caption{Apptronik's Astro was teleoperated to perform a box packing task and a coordinated two-handed box pickup and handover to a human}
\label{fig:capabilities} 
\end{figure}

\section{EXAMPLE CAPABILITY DEMONSTRATION}
\label{sec:demos}
The discussed approach was deployed on the Apptronik Astro robot which has 17 degrees of freedom: two for the torso, three for the neck, and six for each arm. Fig.~\ref{fig:capabilities} shows Astro being teleoperated to perform box packing and handover tasks. Using a similar VR interface from \cite{jorgensen2022cockpit}, a mixed-reality view of the world is given to the user which provides an overlaid preview of IK solutions on top of the current state of the robot to aid with teleoperation in first or third person views. The operator can \textit{clutch} \cite{naceri2021vicarios} a set of joints to command and cycle through different grasp types to perform variable power and pinch grasps with a joystick. The operator can also press a button to maintain the current offset between the hands enabling a coordinated, semi-supervised two-handed box pickup and handover to a human. 

\section{DISCUSSION AND CONCLUSIONS}
Within Apptronik, several individuals with minimal VR and teleoperation experience have successfully performed pick-and-place tasks in our table-top environment requiring only a few minutes to explain how trackers map to robot joints. The inclusion of self-collision avoidance as part of the IK formulation enhanced operational safety and responsiveness as potential collisions were automatically resolved by the IK solver.
Our ongoing hypothesis is that proper allocation of tracker DoFs to joint mapping contributes to the overall intuitiveness of direct teleoperation. Modifying task Jacobians by removing unwanted joint contribution is a simple approach to algorithmically map tracker DoFs to robot joints.

\section{ACKNOWLEDGEMENTS}
The authors would like to acknowledge the supporting personnel at Apptronik that provided full-stack operational support and upkeep of Astro.

\end{document}